%% file: main.tex
\documentclass[10pt, a4paper]{article}
\usepackage{lrec}
\usepackage{graphicx}
\usepackage{tabularx}
\usepackage{soul}

\usepackage{epstopdf}
\usepackage[latin1]{inputenc}

\usepackage{hyperref}
\usepackage{xstring}

\usepackage{comment}
\usepackage{todonotes}
\usepackage{enumitem}

\title{Examining a hate speech corpus\\ for hate speech detection and popularity prediction}

\name{Filip Klubi\v{c}ka, Raquel Fern\'{a}ndez}

\address{School of Computing,
		Dublin Institute of Technology \\
        Institute of Logic, Language and Computation, University of Amsterdam \\
         filip.klubicka@mydit.ie\\
         raquel.fernandez@uva.nl\\}

\abstract{
As research on hate speech becomes more and more relevant every day, most of it is still focused on hate speech detection. By attempting to replicate a hate speech detection experiment performed on an existing Twitter corpus annotated for hate speech, we highlight some issues that arise from doing research in the field of hate speech, which is essentially still in its infancy. We take a critical look at the training corpus in order to understand its biases, while also using it to venture beyond hate speech detection and investigate whether it can be used to shed light on other facets of research, such as popularity of hate tweets.
\\ \newline 
 }

\begin{document}

\maketitleabstract

\section{Introduction}

\input{intro}

\section{Replication: hate speech detection results}
\label{sec:detection}
\input{detection}

\section{New experiment: popularity prediction}
\label{sec:popularity}

\input{popularity}

\section{Corpus analysis}
\label{sec:data}

\input{data}

\section{Conclusion}

This paper has provided an overview of several research directions involving hate speech:

\begin{enumerate}[leftmargin=13pt,itemsep=0pt]
\item A critical look at a publicly available hate speech dataset. 
\item An attempt at replicating and confirming already established hate speech detection findings.
\item Pushing the research space in a new direction: popularity prediction.
\end{enumerate}

Overall, we analyzed a currently popular hate speech dataset, pointed out considerations that have to be made while working such data, and observed that it is biased on several levels. This does not render it useless, but it is important to keep these biases in mind while using this resource and while drawing any sort of conclusions from the data.

As far as replicability goes, the resource does allow one to model hate speech (as biased as it may be), but not without a certain degree of difficulty. We achieve system evaluation scores of 0.71 in terms of F1 score, which is slightly lower than the original results of 0.74 F1 score on the same setup. The differences and gaps in implementation showcase a common trend in scientific publishing - the general problem of reproducing results due to publications not providing sufficient information to make the experiments they describe replicable without involving guessing games. And only when attempting to reproduce a study can one truly realize how much detail is so easily omitted or overlooked, simply due to lack of awareness.

When it comes to popularity prediction, we determine that hate speech negatively impacts the likelihood of likes and replies, but does not affect likelihood of retweets. However, training only on the hate speech portion of the data does seem to boost our model's performance in retweet prediction. These findings, as well as the evaluation scores and feature analyses, are only the first stepping stone in a long line of future work that can be done to better understand the impact of hate speech on social media and how it spreads.

Possibilities include employing social graph mining and network analysis, perhaps using user centrality measures as features in both hate speech and popularity prediction tasks.
In addition, reframing the task as not just a binary prediction task, but rather fitting a regression model to predict the exact number of likes, retweets and replies, would certainly be preferable and more informative, and could lead to a better understanding of how hate speech behaves on Twitter.

What is clear is that hate speech is a very nuanced phenomenon and we are far from knowing everything there is to know about it. Resources are scarce and far from perfect, and much more work and careful consideration are needed, as well as much cleaning, fine-tuning, discussion and agreement on what hate speech even is, if we are to build better resources and successfully model and predict hate speech, or any of its aspects.


\section*{Acknowledgements}

This research was co-funded by the Erasmus+ programme of the European Union and conducted while the first author was visiting the ILLC in Amsterdam. In addition, the research was supported by the ADAPT Centre for Digital Content Technology which is funded under the SFI Research Centres Programme (Grant 13/RC/2106) and is co-funded under the European Regional Development Fund.




\section{Bibliographical References}
\label{main:ref}

\bibliographystyle{lrec}
\bibliography{xample}


\end{document}

%% file: intro.tex
The Internet, likely one of humanity's greatest inventions, facilitates the sharing of ideas and knowledge, as well as online discussion and user interaction. All these are positive features but, as with any tool, whether they are used in a positive or negative manner depends largely on the people that use them. Consequently, and especially when user anonymity is added to the mix, online discussion environments can become abusive, hateful and toxic. 
To help identify, study, and ultimately curb this problem, such negative environments and the language used within are being studied under the name \emph{hate speech}. 

Research on hate speech has become quite prominent in recent years, with dedicated workshops and conferences,\footnote{A few recent examples: {\scriptsize \url{https://europa.eu/newsroom/events/conference-online-hate-speech_en}\\  \mbox{\url{https://sites.google.com/site/abusivelanguageworkshop2017}}\\  \url{http://reportinghate.eu/contact2017/}\\  \url{http://likestiltnorden2017.regjeringen.no/language/en/nordic-hate-speech-conference/}}} and even being featured on LREC2018's list of hot topics. However, hate speech research is still in its infancy. 
In part, this is due to the following challenges: 

\begin{enumerate}[leftmargin=13pt]
\item The term hate speech is difficult to define. 
\newcite{silva16} say that ``hate speech lies in a complex nexus with freedom of expression, group rights, as well as concepts of dignity, liberty, and equality. For this reason, any objective definition (i.e., that can be easily implemented in a computer program) can be contested." 
Generally, the current consensus among researchers seems to be that hate speech can be seen as a phenomenon encompassing issues such as: personal attacks, attacks on a specific group or minority, and abusive language targeting specific group characteristics (e.g., ethnicity, religion, gender, sexual orientation).

\item Creating resources for studying hate speech is far from trivial. Hate speech comprises a very small fraction of online content, and on most social platforms it is heavily moderated. For example, \newcite{nobata16} report that in their corpus of comments on Yahoo! articles collected between April 2014 and April 2015, the percentage of abusive comments is around 3.4\% on Finance articles and 10.7\% on News. Since the phenomenon is elusive, researchers often use lists of offensive terms to collect datasets with the aim to increase the likelihood of catching instances of hate speech \cite{davidson17,waseemhovy16}. This filtering process, however, has the risk of producing corpora with a variety of biases, which may go undetected.

\item Finally, hate speech is present in user-generated content that is not under the control of the researcher. Social media data is typically collected by public APIs that may lead to inconsistent results. For example, \newcite{gonzalez2014assessing} find that the Twitter Search API yields a smaller dataset than the Stream API when using the same filtering parameters. Furthermore, users might delete their profiles or moderate their own questionable content themselves. Thus, datasets on which research experiments are performed are ephemeral, which makes replication of results very difficult. 

\end{enumerate}

In this paper, we focus on the latter two points. We consider a particular hate speech corpus -- a Twitter corpus collected by \newcite{waseemhovy16}, which has been gaining traction as a resource for training hate speech detection models \cite{waseemhovy16,gamback17,park17} -- and analyse it critically to better understand its usefulness as a hate speech resource. In particular, we make the following contributions: 

\begin{itemize}[leftmargin=13pt]
\item We report the outcome of a reproduction experiment, where we attempt to replicate the results by \newcite{waseemhovy16} on hate speech detection using their Twitter corpus.

\item We use the corpus to study a novel aspect related to hate speech: the popularity of tweets containing hate speech. To this end, we develop models for the task of predicting whether a hate tweet will be interacted with and perform detailed feature analyses.

\item We perform a quantitative and qualitative analysis of the corpus to analyse its possible biases and assess the generality of the results obtained for the hate speech detection and popularity tasks. 
\end{itemize}

%% file: detection.tex
We aim to replicate the results on hate speech detection by \newcite{waseemhovy16} using the hate speech Twitter corpus created by the authors.\footnote{\url{https://github.com/zeerakw/hatespeech}} The dataset is a useful resource as it is one of few freely available corpora for hate speech research; it is manually annotated and distinguishes between two types of hate speech -- sexism and racism -- which allows for more nuanced insight and analysis. Additionally, as a Twitter corpus, it provides opportunity for any type of analysis and feature examination typical for Twitter corpora, such as user and tweet metadata, user interaction, etc.

\subsection{Corpus in numbers}

Here we provide just a brief quantitative overview of the corpus, whereas a more detailed qualitative analysis is presented in Section \ref{sec:data}. The original dataset contains 16907 annotated tweets. However, as is common practice with Twitter corpora, the corpus was only made available as a set of annotated tweet IDs, rather than the tweets themselves. To obtain the actual tweets and corresponding metadata, we used the Tweepy Twitter API wrapper.\footnote{\url{http://tweepy.readthedocs.io/en/v3.5.0/}}
Given that the corpus was initially collected and annotated in 2016, there have been some changes in the availability of tweets by the time we extracted in in May 2017. Table \ref{t:annotations} presents the distribution of annotations in the corpus in its original version and the version that was used for this paper. A tweet in the corpus can have three labels (None, Racism, Sexism). It is possible that a tweet has multiple labels, in the case that it contains both racism and sexism (this only happens in 8 tweets in the original dataset, so it is not a widespread phenomenon in this corpus.)

\begin{table}[htbp]
\begin{center}
\begin{tabular}{@{}lrrrr@{}}
\hline
\bf Tag & \bf Original & \bf Available & \bf Deleted & \bf Percent \\
\hline 
None & 11,559 & 11,104 & 455 & 3.94\% \\
Hate & 5,340 & 5,068 & 222 & 4.16\% \\ 
\hspace*{1em}Racism & 1,970 & 1,942 & 22 & 1.12\% \\
\hspace*{1em}Sexism & 3,378 & 3,126 & 200 & 5.92\% \\
\hline
Total & 16,907 & 16,172 & 735 & 4.35\% \\
\hline 
\end{tabular}
\caption{Distribution of hate speech annotations in the corpus. Presenting original counts, available counts, the number of unobtainable tweets and the percentage they represent in their respective category.
 \label{t:annotations}}
\end{center}
\end{table}

The dataset is quite unbalanced, but this is reflective of the unbalanced distribution of hate speech `in the wild', and speaks to why it is so difficult to do research on hate speech in the first place: it is an elusive phenomenon. 
This, combined with the fact that users might delete their profiles or moderate their own questionable content themselves, makes available data scarce, and makes every Twitter corpus smaller over time, and consequently, less valuable and more prone to mistakes when attempting a replicative study.

\subsection{Experimental setup}

As with any replication study, our aim here is to mimic the original experimental setup as closely as possible, in hopes of obtaining same or comparable results. Unfortunately, this effort is already potentially hindered by the fact that the Twitter corpus has shrunk over time. However, the difference is not too large, and we expect it not to have a significant impact on the results.

A much more prominent obstacle is the lack of certain implementation details in the original paper that make reproduction difficult. At several points in the pipeline, we were left to our own devices and resort to making educated guesses as to what may have been done, due to the lack of comprehensive documentation. More specifically, there are two important aspects of the pipeline that present us with this problem: the algorithm and the features.

\paragraph{The algorithm.} \newcite{waseemhovy16} state that they use a logistic regression classifier for their hate speech prediction task. What is not mentioned is which implementation of the algorithm is used, how the model was fit to the data, whether the features were scaled, and whether any other additional parameters had been used.

Due to its popularity and accessibility, we opt for the Scikit-learn \cite{scikit-learn} Python implementation of the logistic regression algorithm.\footnote{\url{http://scikit-learn.org/stable/modules/generated/sklearn.linear_model.LogisticRegression.html}} In addition, after fitting the model, we do not do additional scaling of the features when working with just n-grams (as these are already scaled when extracted), but we do scale our other features using the scaling function.\footnote{\url{http://scikit-learn.org/stable/modules/generated/sklearn.preprocessing.scale.html}} 

\paragraph{The features.} \newcite{waseemhovy16} explore several feature types: they employ n-gram features -- specifically, they find that character n-grams of lengths up to 4 perform best -- and in addition, they combine them with gender information, geographic location information and tweet length, finding that combining n-gram features with gender features yields slightly better results than just n-gram features do, while mixing in any of the other features results in slightly lower scores.


As a rule of thumb, we would attempt to replicate the best performing setup (character n-grams in combination with gender). However, this proved to be difficult, as user gender information is not provided by Twitter (hence it cannot be scraped from the Twitter API) and has not been made available by the authors along with their dataset. 
However, they do describe how they went about procuring the gender information for themselves (by performing semi-automatic, heuristics-based annotation), but only managed to annotate about 52\% of the users. 
This, in combination with the fact that in the original experiment the F1 score improvement when gender is considered is minor (0.04 points) and not statistically significant, led us to focus our efforts on replicating only the experiments involving n-gram features.

However, extracting the n-gram features is also shown to be a nontrivial task, as the original paper does not state how the features are encoded: whether it is using a bag-of-n-grams approach, a frequency count approach, or a TF-IDF measure for each n-gram. We opt for TF-IDF because it is most informative, and just as easy to implement as the more basic approaches.\footnote{\url{http://scikit-learn.org/stable/modules/generated/sklearn.feature_extraction.text.TfidfVectorizer.html}} 



\subsection{Evaluation and results}


The original paper states the use of 10-fold cross-validation for model evaluation purposes, without specifying a particular implementation. For the sake of consistency, we again opt for the Scikit-learn implementation.\footnote{\url{http://scikit-learn.org/stable/modules/generated/sklearn.model_selection.cross_val_score.html}}

We compare the results of our setup to the results of the original experiment. In addition, we also compare evaluations of a system trained on various other features (which we will describe in Section \ref{sec:popularity}) extracted from the tweets and their metadata. The results are presented in Table \ref{t:hs_eval}.

\begin{table}[htbp]
\begin{center}
\begin{tabular}{@{}l|cc|cc|cc@{}}
\hline
\bf Features & \multicolumn{2}{c|}{\bf Original} & \multicolumn{2}{c|}{\bf $n$-grams} & \multicolumn{2}{c}{\bf Other}\\
& Acc & F1 & Acc & F1 & Acc & F1 \\
\hline
\bf Regression & - & 0.74 & 0.84 & 0.71 & 0.79 & 0.65 \\
\hline
\end{tabular}
\caption{Average evaluation scores on the hate speech detection task. The original study only provided an F1 score metric for the logistic regression classifier trained on character $n$-grams (second column). We replicate this experiment (third column), and also train a logistic regression classifier on the same task (fourth column), but on a different set of features (detailed in Section \ref{sec:popularity}). 
 \label{t:hs_eval}}
\end{center}
\end{table}

Examining the table reveals that our best attempt at replicating the original experiment, with logistic regression trained only on character n-grams, yields an F1-score that is 0.03 points lower than the original. Such a drop is to be expected, considering that our version of the dataset was smaller and that we had to fill in some gaps in the procedure ourselves, likely resulting in slight procedural mismatches. However, the drop is not large, and might indicate a stable, consistent result.

When looking at the performance of classifiers trained on features extracted from tweets and their metadata, they significantly underperform, with a 6 point drop compared to our replicated experiment, and a 9 point drop compared to the original results. This adds a strong confirmation of an observation made in the original study, namely that n-gram features are the most predictive compared to any other types of features.

%% file: popularity.tex
To date, most research on hate speech within the NLP community has been done in the area of automatic detection using a variety of techniques, from lists of prominent keywords \cite{warner12} to regression classifiers as seen in the previous section \cite{nobata16,waseemhovy16}, naive Bayes, decision trees, random forests, and linear SVMs \cite{davidson17}, as well as deep learning models with convolutional neural networks \cite{gamback17,park17}.

Our intent in this section is to explore hate speech beyond just detection, using the Twitter corpus by \newcite{waseemhovy16}. Given that Twitter is a platform that enables sharing ideas, and given that extreme ideas have a tendency to intensely spread through social networks \cite{brady17}, our question is: how does the fact that a tweet is a hate tweet affect its popularity?

\subsection{Related work}

To our knowledge there has not been any work relating tweet popularity with hate speech. However, there is a significant body of work dealing with tweet popularity modeling and prediction. Many papers explore features that lead to retweeting. \newcite{suh10} perform an extensive analysis of features that affect retweetability, singling out two groups of features: content and contextual features. 
Similarly, \newcite{zhang12} train a model to predict the number of retweets using two types of features: user features and tweet features. They also compute information gain scores for their features and build a feature-weighted model. They compare the performance of two algorithms: logistic regression and SVM and find that SVM works better, yielding an F-score of 0.71.
In addition, some of the related work also relies on temporal features: \newcite{zaman13} predict the total number of retweets a given amount of time after posting, using a Bayesian model based on features of early retweet times and follower graphs. Similarly, \newcite{hong11} predict the number of retweets, using binary and multi-class classifiers. They use a more varied set of features, and aside from temporal features, they use content, topical and graph features, as well as user metadata.

We do not have temporal data at our disposal, nor are we at this stage interested in predicting the exact number of retweets at any given point. We are more concerned with investigating how hate speech comes into play regarding tweet popularity, if at all.

\subsection{Popularity analysis}

As surveyed above, most of the related work on tweet popularity focuses solely on retweets as indicators of popularity. However, while this is probably the clearest indicator, users can interact with tweets in a number of other ways. For this reason, in the present work we also consider other potential measures of popularity; namely, number of tweet replies and number of `likes' (formerly called `favorites').

The number of likes and retweets in the corpus is varied, but highly skewed, with most of the tweets being liked/retweeted 0 times. The distributions are displayed in Tables \ref{t:interactions-non} and \ref{t:interactions-hate}.

\begin{table}[htbp]
\begin{center}
\begin{tabular}{l|rrrrrrrrr}
\hline
 & \bf 0 & \bf 1 & \bf 2 & \bf 3 & \bf 4 & 5+ \\ 
\hline 
Likes & 9,393 & 1,255 & 246 & 96 & 55 & 59 \\ 
RTs & 10,256 & 755 & 54 & 17 & 9 & 13 \\ 
Replies & 10,304 & 790 & 7 & 3 & 0 & 0 \\ 
\hline
\end{tabular}
\caption{Distribution of the number of interactions on \textbf{non-hate speech} tweets constrained to interactions between users in the corpus. Total number of tweets: 11104.
\label{t:interactions-non}}
\end{center}
\end{table}


\begin{table}[htbp]
\begin{center}
\begin{tabular}{l|rrrrrrrrr}
\hline
 & \bf 0 & \bf 1 & \bf 2 & \bf 3 & \bf 4 & 5+ \\ 
\hline 
Likes & 4,696 & 259 & 49 & 27 & 15 & 22 \\ 
RTs & 4,857 & 180 & 15 & 6 & 3 & 7 \\ 
Replies & 5,049 & 17 & 2 & 0 & 0 & 0 \\ 
\hline
\end{tabular}
\caption{Distribution of the number of interactions on \textbf{hate speech} tweets constrained to interactions between users in the corpus. Total number of tweets: 5068.
\label{t:interactions-hate}}
\end{center}
\end{table}

Given these distributions, we opt for framing the problem as a binary classification task: we wish to determine whether a tweet receives a reaction (retweet, like, response) at least once, or not at all. 

But before we go into prediction, we wish to investigate whether there is a significant difference between hate speech and non-hate speech tweets regarding the number of times a tweet was liked/retweeted/replied to. Thus, to determine whether these differences are statistically significant, we employ the chi-squared ($\chi^2$) statistical significance test.

When examining likes and replies, the test yields $p$-values of \textless0.0001, 
meaning that tweets containing hate speech in the corpus are both liked and replied to significantly less than non-hate speech tweets are. In other words, if a tweet contains hate speech, it is less likely to be liked and replied to. However, when examining the difference in the number of retweets, the $p$-value comes out as 0.5967. This means that we cannot dismiss the null hypothesis, or rather, that whether a tweet contains hate speech or not, does not impact its retweetability either way.

\subsection{Popularity prediction}\label{ss:poppred}


\paragraph{Features.}
We use an large set of features inspired by related work \cite{waseemhovy16,sutton15,suh10,zaman13,hong11,zhang12,cheng14,ma13,zhao15}. We divide our features into three groups: Tweet features (metadata about the the tweet itself), user features (metadata about the author of a tweet) and content features (features derived from the content of the tweet), with the largest number of features falling into the latter group. The features are listed in Table \ref{t:features}. 


\begin{table}[htbp]
\begin{center}
\begin{tabular}{ll}
\hline
\bf Tweet features & \bf User features \\
\hline 
tweet\_age & account\_age \\
tweet\_hour & len\_handle \\
is\_quote\_status & len\_name \\
is\_reply & num\_followers \\
is\_reply\_to\_hate\_tweet & num\_followees \\
num\_replies & num\_times\_user\_was\_listed \\
 & num\_posted\_tweets \\
 & num\_favorited\_tweets \\
\hline
\multicolumn{2}{c}{\bf Content features} \\
\hline 
is\_hate\_tweet & has\_uppercase\_token \\
has\_mentions & uppercase\_token\_ratio \\ 
num\_mentions & lowercase\_token\_ratio \\
has\_hashtags & mixedcase\_token\_ratio \\
num\_hashtags & blacklist\_total \\
has\_urls & blacklist\_ratio \\
num\_urls & total\_negative\_tokens \\
char\_count & negative\_token\_ratio \\
token\_count & total\_positive\_tokens \\
has\_digits & positive\_token\_ratio \\
has\_questionmark & total\_subjective\_tokens \\
has\_exclamationpoint & subjective\_token\_ratio \\
has\_fullstop & \\ 
\hline
\end{tabular}
\caption{Features used in the popularity prediction task.
 \label{t:features}}
\end{center}
\end{table}




\paragraph{Models and results.}
We train a logistic regression classifier, as well as a linear SVM classifier to compare their performances. We also train separate models for likes and for retweets. One pair of models was trained on the whole corpus, and two additional pairs of classifiers were trained on just the hate speech portion and non-hate speech portion of the corpus respectively.

We tested all models using 10-fold cross validation, holding out 10\% of the sample for evaluation to help prevent overfitting. All modeling and evaluation was performed using Scikit-learn \cite{scikit-learn}. 
The evaluation results are presented in Table \ref{t:eval}. We also make our feature dataset, and our training and evaluation scripts available to the community for transparency and reproduction purposes.\footnote{The dataset is comprised of anonymized tweet IDs with extracted content features.

Link to GitHub repository: \url{https://github.com/GreenParachute/hate-speech-popularity}. }


Interestingly, our classifiers are consistently better at predicting retweets than likes.
Given that they are trained on the same features, this indicates that the nature of these two activities is different, in spite of the fact that they intuitively seem very similar.

Furthermore, it seems that the linear regression model seems to perform slightly better overall than the SVM model on both prediction tasks (likes and retweets).

\begin{table}[htbp]
\begin{center}
\begin{tabular}{@{}l|cc|cc|cc@{}}
\hline
 & \multicolumn{2}{c|}{\bf Whole dataset} & \multicolumn{2}{c|}{\bf Non-hate} & \multicolumn{2}{c}{\bf Hate}\\
& Acc & F1 & Acc & F1 & Acc & F1 \\
\hline
Regression \\
\hline
\bf Likes & 0.75 & 0.57 & 0.73 & 0.63 & 0.79 & 0.29 \\
\bf Retweets & 0.82 & 0.69 & 0.81 & 0.68 & 0.85 & 0.73 \\
\hline
SVM \\
\hline
\bf Likes & 0.74 & 0.58 & 0.73 & 0.63 & 0.79 & 0.16 \\
\bf Retweets & 0.81 & 0.66 & 0.82 & 0.69 & 0.84 & 0.70 \\
\hline
\end{tabular}
\caption{Average evaluation scores on binary prediction task (predicting if a tweet will be liked/retweeted or not). Presenting results with different subsets of the corpus and comparing performance of logistic regression and SVM models.
 \label{t:eval}}
\end{center}
\end{table}

\paragraph{Analysis.}
In order to investigate which features are most informative for the task, we perform feature ablation according to our feature groups. Some notable results show that removing author metadata from the feature set reduces the performance of the model.\footnote{As our analysis in Section~\ref{sec:data} will reveal, this seems a consequence of a strong bias towards a handful of overproductive authors in the corpus.} However, the biggest takeaway for now is the $is\_reply$ feature's impact on the model. Our SVM model's average accuracy drops by 0.04 points if the $is\_reply$ feature is omitted from the feature set, whereas omitting many of the other features decreases performance scores by 0.02 points at most, if at all. 

Inspired by \newcite{zhang12}, we also calculate information gain for all features. The top most informative features for each task (predicting whether a tweet will be liked/retweeted) and for each setup (full dataset/non-hate dataset/hate dataset) according to the information gain (IG) measure are presented in Tables \ref{t:IG1}, \ref{t:IG2} and \ref{t:IG3}.

\begin{table}[htbp]
\begin{center}
\begin{tabular}{@{}lr|lr@{}}\hline
\multicolumn{2}{c|}{\bf Liking} & \multicolumn{2}{c}{\bf Retweeting} \\
Feature & IG & Feature & IG \\
\hline
num\_tweets & 0.1005 & is\_reply & 0.0811 \\
num\_followees & 0.0999 & uppercase\_ratio & 0.0541 \\
num\_liked\_tweets & 0.0992 & has\_uppercase & 0.0517 \\
num\_followers & 0.0985 & char\_count & 0.0509 \\
user\_id & 0.0985 & num\_tweets & 0.0334 \\ 
account\_age & 0.098 & num\_liked\_tweets & 0.0329 \\
num\_listed & 0.0954 & num\_followees & 0.0323 \\
len\_name & 0.0732 & num\_followers & 0.0307 \\ 
\hline
\end{tabular}
\caption{Most informative features according to information gain (IG) scores for the \textbf{whole dataset}. (A higher score indicates bigger importance.)
 \label{t:IG1}}
\end{center}
\end{table}

\begin{table}[htbp]
\begin{center}
\begin{tabular}{@{}lr|lr@{}}\hline
\multicolumn{2}{c|}{\bf Liking} & \multicolumn{2}{c}{\bf Retweeting} \\
Feature & IG & Feature & IG \\
\hline
num\_followees & 0.0648 & is\_reply & 0.1155\\
num\_followers & 0.0623 & uppercase\_ratio & 0.0876 \\
num\_tweets & 0.0622 & has\_uppercase & 0.0877 \\
account\_age & 0.0605 & num\_liked\_tweets & 0.0532 \\
user\_id & 0.0605 & num\_listed & 0.0529  \\
num\_listed & 0.0595 & num\_followers & 0.0524 \\
len\_handle & 0.0552 & num\_tweets & 0.0511 \\
len\_name & 0.0515 & num\_followees & 0.0495 \\
\hline
\end{tabular}
\caption{Most informative features according to information gain (IG) scores for the \textbf{hate speech subset}. (A higher score indicates bigger importance.)
 \label{t:IG2}}
\end{center}
\end{table}

\begin{table}[htbp]
\begin{center}
\begin{tabular}{@{}lr|lr@{}}\hline
\multicolumn{2}{c|}{\bf Liking} & \multicolumn{2}{c}{\bf Retweeting} \\
Feature & IG & Feature & IG \\
\hline
num\_followers & 0.0974 & is\_reply & 0.0677 \\
num\_followees & 0.0970 & char\_count & 0.0519\\
user\_id & 0.0963 & uppercase\_ratio & 0.042\\
account\_age & 0.0963 & has\_uppercase & 0.0396 \\
num\_tweets & 0.0953 & token\_count & 0.0323\\
num\_liked\_tweets & 0.0948 & num\_followees & 0.0258\\
num\_listed & 0.0941 & num\_liked\_tweets & 0.0257\\
len\_name & 0.077 & num\_followers & 0.0246\\
\hline
\end{tabular}
\caption{Most informative features according to information gain (IG) scores for the \textbf{non-hate speech subset}. (A higher score indicates bigger importance.)
 \label{t:IG3}}
\end{center}
\end{table}

Given the context of this paper and the nature of the corpus, it is interesting to note that the $is\_hate\_tweet$ feature does not appear anywhere near the top of the IG rankings, indicating that it is not very informative in regards to predicting whether a tweet will be liked or retweeted. 

On a broader note, although the feature lists are more or less similar across the different dataset splits, there is a marked difference between the retweeting and liking lists, in each split. Features that are very informative for retweeting, but not for liking, are whether the tweet contains uppercase tokens, and, most notably, whether the tweet is a reply. This is in line with our findings in the feature ablation study, confirming that there is a strong link between the possibility of retweeting and whether or not the tweet in question is a reply. Our interpretation of this discrepancy is that original, stand-alone ideas (tweets) might be more likely to be picked up and passed on (retweeted), than a turn in a twitter conversation thread would be. In addition, these overall IG measurements also indicate that there is an inherent qualitative difference between the acts of liking and retweeting. 

%% file: data.tex
As the field of hate speech research is yet to mature, with disagreement about what exactly the phenomenon entails \cite{Waseem17} and without a unified annotation framework \cite{Fiser17}, it is warranted to look at the data and examples in more detail, with considerations for potential shortcomings. In Section~\ref{sec:detection}, we pointed out the ephemeral nature of the corpus by \newcite{waseemhovy16}, common to all Twitter datasets. In this section, we analyse other characteristics of the corpus related to the challenges of data collection for hate speech analysis we mentioned in the Introduction (point 2), which can result in undesirable biases.

\paragraph{Tweet collection.} 
Given the small fraction of online content comprised of hate speech, collecting a significant amount of examples is an extremely difficult task. At present, it is not feasible to collect a large sample of tweets and then manually label them as hate or non hate, as the fraction of instances labeled with the positive class will be negligible. The only way to model the phenomenon is to target tweets already likely to contain hate speech. 

Driven by this rationale, the authors of the corpus have obtained their dataset by performing an initial manual search of common slurs and terms used pertaining to religious, sexual, gender, and ethnic minorities. The full list of terms they queried for is not very long: \textit{MKR}, \textit{asian drive}, \textit{femi-nazi}, \textit{immigrant}, \textit{nigger}, \textit{sjw}, \textit{WomenAgainstFeminism}, \textit{blameonenotall}, \textit{islam terrorism}, \textit{notallmen}, \textit{victimcard}, \textit{victim card}, \textit{arab terror}, \textit{gamergate}, \textit{jsil}, \textit{racecard}, \textit{race card}. In the results obtained from these queries, they identified frequently occurring terms in tweets that contain hate speech and references to specific entities (such as \textit{MKR}, addressed further below). In addition to this, they identified a small number of prolific users from these searches.

This manner of tweet collection allowed the authors to obtain quite a considerable amount of data. However, this approach to data collection inevitably introduces many biases into the dataset, as will be demonstrated further in this section.

\paragraph{Qualitative observations on tweet content.}
According to the annotation guidelines devised by \newcite{waseemhovy16} for the purpose of annotating this corpus, a tweet is tagged as offensive if it:
(1) uses a sexist or racial slur, (2) attacks a minority, (3) seeks to silence a minority, (4) criticizes a minority (without a well founded argument), (5) promotes, but does not directly use, hate speech or violent crime, (6) criticizes a minority and uses a straw man argument, (7) blatantly misrepresents truth or seeks to distort views on a minority with unfounded claims, (8) shows support of problematic hashtags (e.g. \#BanIslam, \#whoriental, \#whitegenocide), (9) negatively stereotypes a minority,
(10) defends xenophobia or sexism, (11) the tweet is ambiguous (at best); and contains a screen name that is offensive as per the previous criteria; and is on a topic that satisfies any of the above criteria.

Though at first glance specific and detailed, these criteria are quite broad and open to interpretation. This was likely done to cover as many hate speech examples as possible -- a thankless task, as hate speech data is scarce to begin with. However, due to this same breadth, the corpus contains some potential false positives. The most jarring example of this being that, if a user quotes a tweet containing hate speech (by prepending the quoted text with ``RT"), the quoter's tweet is tagged as hate speech. 
Certainly, the user could have quoted the original tweet in support of its message, and even if not, one could argue that they do perpetuate the original hateful message by quoting it. On the other hand, it is just as likely that the user is quoting the tweet not to make an endorsement, but a neutral response. It is even more likely that the user's response is an instance of \emph{counterspeech} --- interaction used to challenge hate speech \cite{Wright17}. Manual inspection shows that there are instances of both such phenomena in the corpus, yet all those tweets are tagged as hate speech. In fact, $\sim$30\% of hate speech tweets in the corpus contain the token 'RT', indicating they are actually retweets. This could pose a problem further down the line when extrapolating information about hate speech users. Addressing this issue would at the very least require going through tweets with quotes and relabeling them, if not altogether rethinking the annotation guidelines, or rather, being more mindful of the semantics at play during annotation.

\paragraph{Topic domain.} 
In spite of the broad guidelines, however, it seems that the actual hate speech examples end up falling on quite a narrow spectrum. Even though the tweets were semi-automatically picked based on a wide variety of keywords likely to identify hate speech, the tag `racism' is in fact used as an umbrella term to label not only hate based on race/ethnicity, but also religion, specifically Islam. Indeed, the majority of the tweets tagged as racist are, in fact, islamophobic, and primarily written by a user with an anti-Islam handle (as per guideline 11). Though it is stated in the original paper which seed words were used to collect the data (which included both racist and islamophobic terms), it is undeniable that the most frequent words in the racist portion of the corpus refer to islamophobia (which is also explicitly stated by the authors themselves). This is not wrong, of course, but it begs the question of why the authors did not choose a more specific descriptor for the category, especially given that the term `racism' typically sparks different connotations, ones that, in this case, do not accurately reflect the content of the actual data.

When it comes to sexist tweets, they are somewhat more varied than those annotated as racist. However, they do contain a similar type of bias: $\sim$13.6\% of the tweets tagged as sexist contain the hashtag and/or handle MKR/MyKitchenRules. \emph{My Kitchen Rules} is an Australian competitive cooking game show which is viewed less for the cooking and skill side of the show than for the gossip and conflict which certain contestants are encouraged to cause.\footnote{\url{https://en.wikipedia.org/wiki/My_Kitchen_Rules\#Criticism}} It seems to be a popular discussion topic among fans of the show on Twitter, and apparently prompts users to make sexist remarks regarding the female contestants. There is nothing inherently problematic about this being included in a corpus of hate speech, but it cannot be disregarded that more than a tenth of the data on sexism is constrained to an extremely specific topic domain, which might not make for the most representative example of sexism on Twitter.

\paragraph{Distribution of users vs. tweet content}
Another interesting dimension of the corpus that we observe is the distribution of users in relation to the hate speech annotations --  an aspect that could be important for our analysis of popularity presented in Section~\ref{sec:popularity}

There are 1858 unique user IDs in the corpus. Thus many of the 16907 tweets were written by the same people. As a simplistic approximation, we can (very tentatively) label every user that is the author of at least one tweet containing hate speech as a hate user; and users that, in the given dataset, have not produced any tweets containing hate speech we label as non-hate users. Of course, this does have certain drawbacks, as we cannot know that a user does not produce hate speech outside the sample we are working with, but it does provide at least an approximation of a user's production of hate tweets in the sample. Using this heuristic, the distribution of users in the corpus in regards to whether they produce hate speech or not is presented in Table \ref{t:users}.

\begin{table}[htbp]
\begin{center}
\begin{tabular}{lr}
\hline
\bf User type & \bf Count \\
\hline 
Non-hate & 1,334 \\
Hate & 525 \\ 
\hspace*{1em}Racist & 2 \\
\hspace*{1em}Sexist & 520 \\
\hspace*{1em}Racist and sexist & 3 \\
\hline 
Total & 1,859 \\
\hline
\end{tabular}
\caption{Distribution of users according to the type of tweets they produce.
 \label{t:users}}
\end{center}
\end{table}



\begin{figure}[h!]
	\centering
  \hspace*{-7pt}  
	\includegraphics[width=1.03\columnwidth]{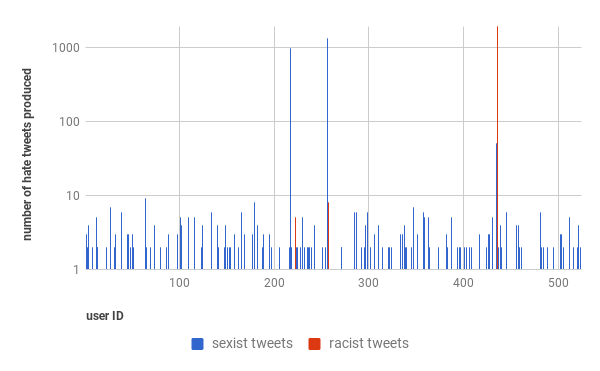}
\vspace*{-25pt}
	\caption{Graph illustrating the distribution of tweets containing hate speech among users producing them. We represent the number of tweets in logarithmic scale.
    \label{f:hate-distr}}
\end{figure}

A really striking discrepancy immediately jumps out when looking at Table \ref{t:users}: there is a total of 5 users responsible for the 1942 racist tweets, as opposed to the 523 users responsible for the 3126 sexist tweets. Assuming normal distribution (which is certainly the wrong assumption), on average there are 388 racist tweets per racist user, while there is an average of 6 sexist tweets per sexist user. The actual distribution, however, is extremely skewed -- the bulk of all the hate speech data is distributed between three users: one user who produced 1927 tweets tagged as racist, and two users who respectively produced 1320 and 964 tweets tagged as sexist. This is illustrated in Figure \ref{f:hate-distr}.

Such a distribution renders any attempt at generalization or modeling of racist tweets moot, as the sample cannot be called representative of racism as such, but only of the Twitter production of these 5 users.\footnote{However, the data might still be useful when looked at in bulk with sexism, as it might reinforce the similarities they both share stemming from the fact that they are types of hate speech.} Similarly, the fact that most of the tweets tagged as sexist belong to the same two users considerably skews this subset of the data. 

\paragraph{Corollary.}
All of these points deserve due consideration. The imbalances with respect to distribution of users were certainly considered while we worked with the data. In an attempt to reduce them, we did not distinguish between racist and sexist tweets in our analysis in both Sections \ref{sec:detection} and \ref{sec:popularity} (even though we were tempted to do so), but rather treated them all as simply hate speech tweets. 
Additionally, it is possible that the insights and biases presented in this section might even call into question the relevance of the findings from Section \ref{sec:popularity}, as the popularity modeled there is likely reflecting the popularity of the particular Twitter users in the corpus rather than of hate speech tweets as such.